\documentclass[11pt]{article}

\usepackage{microtype}
\usepackage{graphicx}
\usepackage{subfigure}
\usepackage{booktabs} 

\usepackage{algorithm,algorithmic}

\RequirePackage[authoryear,round]{natbib}
\usepackage{fullpage}
\usepackage[utf8]{inputenc}
\usepackage[T1]{fontenc}
\usepackage{microtype}
\usepackage{csquotes}
\usepackage[osf,sc]{mathpazo}
\RequirePackage[scaled=0.90]{helvet}
\RequirePackage[scaled=0.85]{beramono}
\RequirePackage{textcomp}
\usepackage{xcolor}
\definecolor{DarkGreen}{rgb}{0.075,0.375,0.075}
\definecolor{DarkRed}{rgb}{0.5,0.1,0.1}
\definecolor{DarkBlue}{rgb}{0.1,0.1,0.5}
\definecolor{Gray}{rgb}{0.2,0.2,0.2}
\usepackage[pdftex]{hyperref}
\hypersetup{
    unicode=false,          
    pdftoolbar=true,        
    pdfmenubar=true,        
    pdffitwindow=false,      
    pdfnewwindow=true,      
    colorlinks=true,       
    linkcolor=DarkBlue,          
    citecolor=DarkGreen,        
    filecolor=DarkRed,      
    urlcolor=DarkBlue,          
    pdftitle={},
    pdfauthor={},
}
\usepackage[skip=2mm,indent=5mm]{parskip}
\usepackage[small]{caption}
\RequirePackage{amsthm,amsmath,amsfonts,amssymb}

\usepackage[capitalize,noabbrev]{cleveref}

\theoremstyle{plain}
\newtheorem{theorem}{Theorem}[section]

\newtheorem{lemma}[theorem]{Lemma}

\theoremstyle{definition}

\theoremstyle{remark}

\usepackage[textsize=tiny]{todonotes}

\usepackage{pifont}
\usepackage{hyperref}
\usepackage{graphicx} 
\usepackage{booktabs} 

\usepackage{epsfig}
\usepackage{subfigure}
\usepackage[utf8]{inputenc}

\usepackage[T1]{fontenc}

\usepackage{amsmath}
\usepackage{subfigure}
\usepackage{amsfonts}
\usepackage{amsthm}
\usepackage{amssymb}
\usepackage{pifont}
\usepackage{multirow}
\usepackage{colortbl}
\usepackage{nameref}
\usepackage{bm}
\usepackage{enumitem}

\def\eqref#1{equation~\ref{#1}}

\def\1{\bm{1}}

\def\vtheta{{\bm{\theta}}}

\def\vr{{\bm{r}}}
\def\vs{{\bm{s}}}

\def\mL{{\bm{L}}}

\def\mR{{\bm{R}}}
\def\mS{{\bm{S}}}

\DeclareMathAlphabet{\mathsfit}{\encodingdefault}{\sfdefault}{m}{sl}
\SetMathAlphabet{\mathsfit}{bold}{\encodingdefault}{\sfdefault}{bx}{n}

\def\gG{{\mathcal{G}}}

\def\gL{{\mathcal{L}}}
\def\gM{{\mathcal{M}}}

\def\gT{{\mathcal{T}}}

\newcommand{\ie}{\emph{i.e.}}

\definecolor{DarkRed}{rgb}{0.5,0.1,0.1}

\newtheorem{property}[theorem]{\textbf{Property}}

\title{Inherent Trade-Offs between Diversity and Stability in Multi-Task Benchmarks}
\author{%
  Guanhua Zhang\and Moritz Hardt
}
\date{\textit{Max Planck Institute for Intelligent Systems, Tübingen and Tübingen AI Center}}

\begin{document}

\onecolumn
\maketitle

\begin{abstract}
We examine multi-task benchmarks in machine learning through the lens of social choice theory. We draw an analogy between benchmarks and electoral systems, where models are candidates and tasks are voters.  This suggests a distinction between cardinal and ordinal benchmark systems. The former aggregate numerical scores into one model ranking; the latter aggregate rankings for each task. We apply Arrow's impossibility theorem to ordinal benchmarks to highlight the inherent limitations of ordinal systems, particularly their sensitivity to the inclusion of irrelevant models.  Inspired by Arrow's theorem, we empirically demonstrate a strong trade-off between diversity and sensitivity to irrelevant changes in existing multi-task benchmarks. Our result is based on new quantitative measures of diversity and sensitivity that we introduce. Sensitivity quantifies the impact that irrelevant changes to tasks have on a benchmark. Diversity captures the degree of disagreement in model rankings across tasks.  We develop efficient approximation algorithms for both measures, as exact computation is computationally challenging. Through extensive experiments on seven cardinal benchmarks and eleven ordinal benchmarks, we demonstrate a clear trade-off between diversity and stability: The more diverse a multi-task benchmark, the more sensitive to trivial changes it is. Additionally, we show that the aggregated rankings of existing benchmarks are highly unstable under irrelevant changes.
The codes and data are available at \url{https://socialfoundations.github.io/benchbench/}.
\end{abstract}

\section{Introduction}

At this point, there is little agreement about what the right benchmark is for different tasks in machine learning~\citep{ethayarajh2020utility, bowman2021will, kiela2021dynabench}. Natural language understanding, for example, has hundreds of different benchmarks, each measuring different qualities of a model~\citep{storks2019recent}. On the one hand, multiple benchmarks are desirable when it comes to creating a diverse canvas of evaluation results. On the other hand, the plurality of different benchmarks makes it challenging to consistently measure progress, as different benchmarks suggest different model rankings.

The de facto solution to the problem are multi-task benchmarks~\citep{wang2018glue, wang2019superglue,hendrycks2020measuring}. Major recent developments, such as BigBench~\citep{srivastava2022beyond} and HELM \citep{liang2022holistic}, combine hundreds of evaluation tasks into a single benchmark. The hope is that by aggregating many tasks into one, a reliable and representative picture of model performance will emerge.

In this work, we scrutinize multi-task benchmarks through the lens of social choice theory. In doing so, we analogize multi-task benchmarks with electoral systems. Models stand in analogy with candidates in the electoral system, and tasks with voters. Each task in a multi-task benchmark may rank candidate models differently. The benchmark must determine a ranking of candidates given the different votes.

A robust lesson from social choice theory is that there is no perfect voting system. Celebrated results, such as Arrow's impossibility theorem~\citep{arrow1950difficulty,arrow2012social}, point at inherent limitations in the design of desirable voting rules~\citep{taylor2005social}. Inspired by social choice theory, we surface an important trade-off in multi-task benchmarks between a measure of diversity and a measure of robustness to irrelevant changes. In a nutshell, we demonstrate empirically that current multi-task benchmarks fail to be both robust and diverse. Instead, one comes at the expense of the other.

\begin{figure}
    \centering
    \includegraphics[width=0.95\linewidth]{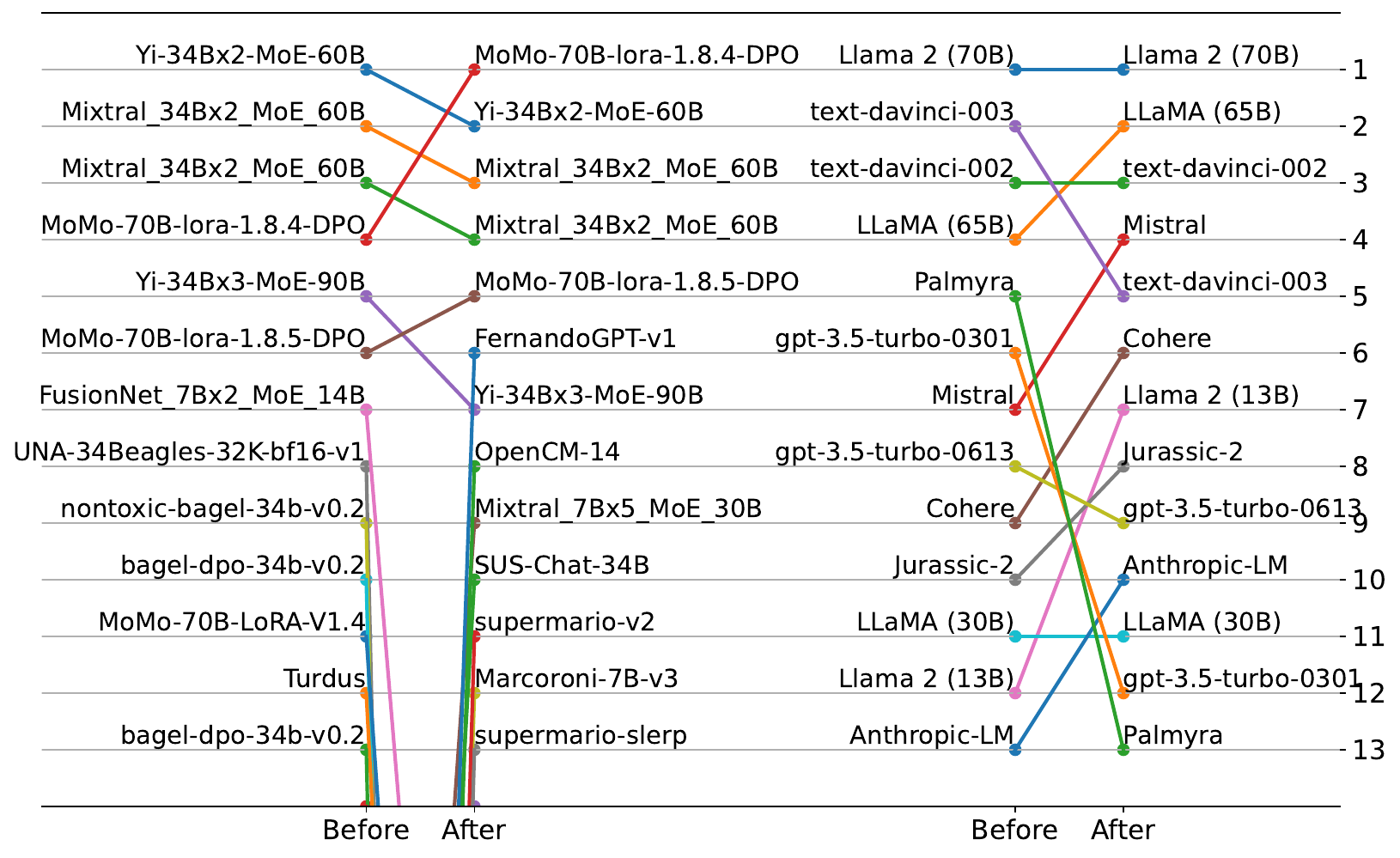}
    \caption{Ranking changes after irrelevant changes on tasks. 
    For cardinal benchmark \texttt{OpenLLM} (left), \textit{Before} refers to the original ranking, and \textit{After} is the new ranking after injecting label noises into different tasks.
    For ordinal benchmark \texttt{HELM-accuracy} (right), 
    \textit{Before} refers to the ranking based on only the original top-20\% models, while \textit{After} is the new relative ranking after adding irrelevant models from the rest 80\%.
    $y$-axis refers to the ranking.
    }
    \label{fig:banner}
    
\end{figure}

\subsection{Our contributions}

We propose a distinction between \emph{cardinal benchmark systems} and \emph{ordinal benchmark systems}. Cardinal benchmark systems aggregate multiple rankings into one on the basis of numerical scores, such as accuracy numbers.  Ordinal benchmark systems instead aggregate rankings into a single ranking. BigBench is an example of a cardinal benchmark, ranking by average accuracy over the tasks. HELM is an example of an ordinal system, comparing any two models by how often one ranks higher than the other.

To start, we point out that Arrow's impossibility result directly applies to ordinal systems, such as HELM. We observe that in the case of ordinal benchmarks, the desirable property that fails in Arrow's language is \emph{independence of irrelevant alternatives}. What this means is that adding irrelevant models to a benchmark can perturb the order of top contenders. We demonstrate that this is indeed possible with HELM and similar benchmarks.

Arrow's result is neither quantitative, nor does it apply to cardinal systems. Inspired by Arrow's theorem, we introduce a quantitative measure of \emph{sensitivity to irrelevant changes}. Sensitivity measures how responsive a benchmark is to trivial transformations of tasks. For example, adding a fraction of random labels to a task does not change the relative performance of models, thus resulting in an equivalent task. 

It is easy to design benchmarks that are robust to such irrelevant changes. Simply take a single-task benchmark. Or take the single task and copy it many times to obtain a multi-task benchmark. We therefore contrast our measure of sensitivity with a measure of \emph{diversity}. Diversity measures the degree to which different tasks disagree in their model rankings.  Multi-task benchmarks lacking in diversity are essentially single-task benchmarks.

Our measures of diversity and sensitivity are computationally hard to compute exactly. We therefore developed efficient approximation algorithms for both.

Through comprehensive experiments, we demonstrate that there is a strong trade-off between diversity and sensitivity in current multi-task benchmarks. The more diverse a multi-task benchmark, the more sensitive to trivial changes it is. 
In other words, the pursuit of diversity compromises sensitivity, and striving for robustness necessitates sacrificing diversity.
We confirm this trade-off in seven cardinal benchmarks and eleven ordinal benchmarks from natural language understanding and computer vision.

\begin{figure}[t!]
    \centering
    \includegraphics[width=0.46\textwidth]{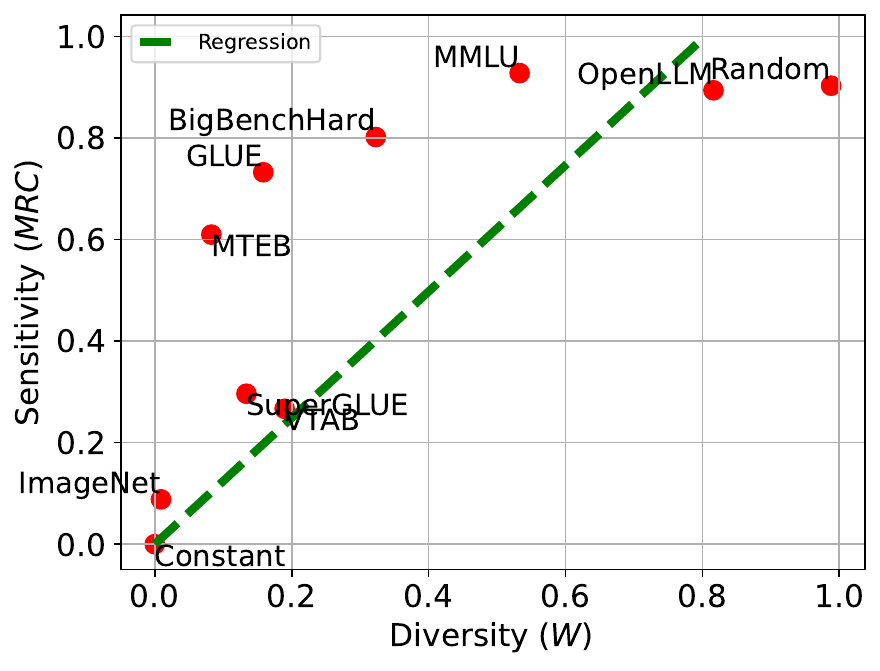}
    \includegraphics[width=0.46\textwidth]{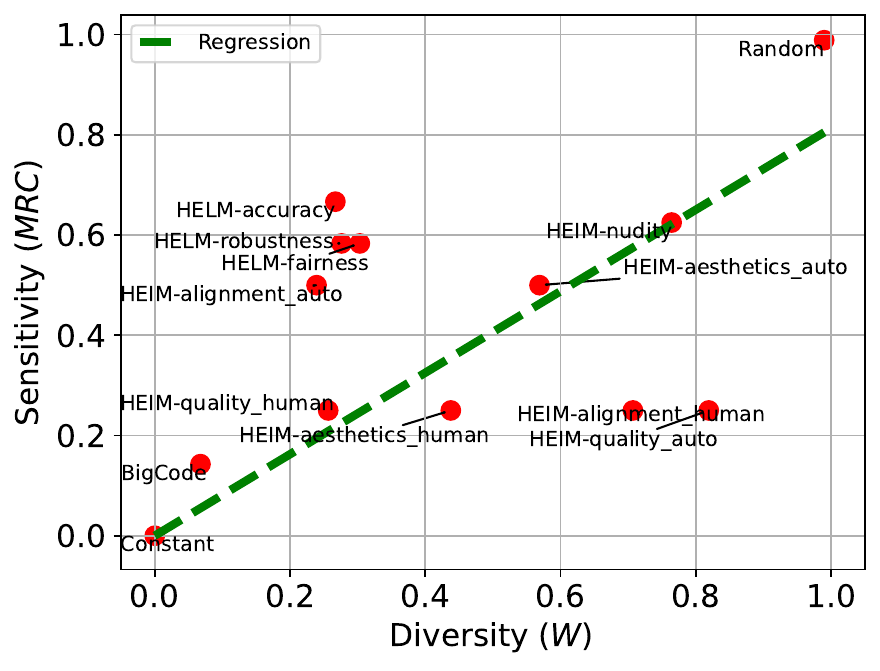}    
    \caption{Trade-off between benchmark diversity and sensitivity to irrelevant changes. Left: Cardinal benchmarks. Right: Ordinal benchmarks. Sensitivity is measured in terms of the maximum normalized rank change (MRC) possible via irrelevant task changes. Diversity is measured by Kendall's coefficient of concordance ($W$). The green curve is a linear regression on all points without fitting the intercept.
    }
    \label{fig:cardinal_trade_off_intro}
\end{figure}

The most stable benchmark, by our measure, is a constant benchmark. The most diverse benchmark is a random benchmark. We show that all existing multi-task benchmarks strike a trade-off no better than a linear interpolation between random and constant. 
In particular, our empirical analysis reveals that current benchmarks are highly unstable to irrelevant changes.
For illustration, Figure~\ref{fig:banner} gives an example where both a cardinal benchmark (\texttt{OpenLLM}) and an ordinal benchmark (\texttt{HELM-accuracy}) suffer from significant ranking changes after trivial task transformations. Figure~\ref{fig:cardinal_trade_off_intro} summarizes the trade-off between diversity and sensitivity for both cardinal and ordinal benchmarks.

\section{Related Works}
Benchmarks are at the foundation of applied machine learning research, underpinning many of its successes~\citep{Donoho2023DataSA, koch2021Reduced, Zhang2019MachineLT,ott2022Mapping}. Although many benchmarks have been proposed, far fewer works have studied benchmarks as a scientific subject itself; see \cite{hardt2022patterns} for an overview. 
With recent machine learning models achieving impressive abilities across many different evaluation settings~\citep{Ramesh2021ZeroShotTG,Anil2023GeminiAF,openai2023GPT4,Touvron2023LLaMAOA,Touvron2023Llama2O}, the spotlight has increasingly turned to multi-task benchmarks.  Multi-task benchmarks aim to provide a diverse and holistic evaluation of machine learning models by covering many different tasks and metrics~\citep{liang2022holistic,Lee2023HolisticEO,srivastava2022beyond,wang2018glue,wang2019superglue}.
Concurrently, various concerns regarding benchmarks have surfaced, highlighting their limitations, see, e.g.,~\citep{liao2021we,bowman2021will,Zhang2023LLMEvalAP,Boubdir2023EloUR}, in particular, susceptibility to chosen tasks~\citep{dehghani2021Benchmark,Alzahrani2024WhenBA}, non-smooth utility functions~\citep{ethayarajh2020utility}, data contamination~\citep{Roberts2023DataCT,Magar2022DataCF}, possibility for overfitting due to repeated use of test sets~\citep{Dwork2014PreservingSV, Blum2015TheLA, Feldman2019TheAO, Mania2019ModelSM, Arora2021RipVW}. \citet{shirali2022theory} demonstrated inherent limitations of \emph{dynamic benchmarks}, another recent benchmark design paradigm that aims to mitigate shortcomings of static single-task benchmarks.

In our research, we focus on the challenge of aggregating performance measures within multi-task benchmarks. 
While existing studies have raised concerns about 
the aggregation problem in benchmarks, they have primarily focused on cardinal aggregation by mean scores~\citep{Mania2019ModelSM,Colombo2021InfoLMAN,peyrard-etal-2017-learning,Mishra2021HowRA}. 
As a result, there is a shift towards ordinal benchmarks where only relative performances in each task are used for aggregation~\cite{liang2022holistic, Lee2023HolisticEO}.
\citet{himmi2023More} adopt a compatible partial ranking approach to address missing scores in benchmarks and introduce a Borda count-based aggregation method \citep{Kelly1988SocialCT}. 
\citet{colombo2022What} propose a new aggregation process to fix the scale difference problem of cardinal based on Kemeny consensus~\citep{Shapiro1993MathematicsWN}.
\citet{rofin2023Vote} propose VOTE’N’RANK, which comprises eight procedures that depend on rankings for each task. 
In our study, we highlight the fundamental compromise one must navigate between diversity and stability for both cardinal and ordinal benchmarks.

\section{A Social Choice Perspective for Benchmarks}
\label{sec:perspective}
Social choice theory addresses the problem of aggregating individual preferences to select the best option or candidate~\citep{Kelly1988SocialCT}. In the context of multi-task machine learning benchmarks, we adopt this framework by considering each task as an individual voter. The tasks, as voters, provide preference scores to different candidate models, akin to how individuals might vote for political candidates. The problem of aggregating these task-based votes into a cohesive ranking of models parallels the challenge in social choice of electing a candidate that best represents the preferences of the electorate.

From this perspective, we divide multi-task benchmarks into two classes: cardinal benchmarks and ordinal benchmarks.
Cardinal benchmarks collect model scores from tasks, translate quantitative performance into a single average score per model, and rank all candidate models based on it.
Ordinal benchmarks, on the other hand, only utilize relative rankings rather than absolute scores. 
Every task ranks the models based on performance, and the final model rankings emerge from an aggregation of these ordinal positions. 

\paragraph{Notation.}
We need some notation to formalize the problem:
\begin{itemize}[itemsep=0mm, topsep=0mm]
    \item $\gT=(T_1, T_2, \ldots, T_n)$ represents the list of all $n$ tasks in the benchmark, analogous to voters.
    \item $\gM$ refers to the set of all potential candidate models that could be evaluated by the benchmark. 
    \item Let $\gL=(L_1, L_2, \ldots, L_m)$ be any non-empty list of candidate models with $m$ models, where $L_i \in \mathcal{M}$ for any $i$.
    \item For any $\gL$, we define $s_{ij}$ as the score for the $i$-th model in $\gL$ in task $T_j$.
    For simplicity, we abuse the notations and use $\vs_j=(s_{1j}, s_{2j}, \ldots, s_{mj})$ as scores in any task $T_j$, and $\mS=(\vs_1, \vs_2, \ldots, \vs_n)$ as scores over all tasks.
    \item For any $\gL$, we define $r_{ij}$ as the rank for the $i$-th model $\gL$ in task $T_j$ \emph{w.r.t.} $\gL$.
    For simplicity, we abuse the notations and use $\vr_j=(r_{1j}, r_{2j}, \ldots, r_{mj})$ as ranks in any task $T_j$, and $\mR=(\vr_1, \vr_2, \ldots, \vr_n)$ as ranks over all tasks.
    \item A cardinal benchmark is defined as a function $f^{\text{c}}=h^{\text{c}} \circ g^{\text{c}}$, which is composed of the scoring function $g^{\text{c}}$ and the aggregation function $h^{\text{c}}$. Specifically, $g^{\text{c}}$ takes a list of models $\gL$ as input and outputs the corresponding scores for each index over all tasks, \ie, $\mS=g^{\text{c}}(\gL)$. 
    The scores $\mS$ are fed into 
    $h^{\text{c}}$, which outputs the final ranking $\vr^{\text{c}}=(r^{\text{c}}_1,r^{\text{c}}_2,\ldots,r^{\text{c}}_m)$, \ie, $\vr^{\text{c}}=h^{\text{c}}(\mS)$.
    \item An ordinal benchmark is defined as a function $f^{\text{o}}=h^{\text{o}} \circ g^{\text{o}}$, which is composed of the scoring function $g^{\text{o}}$ and the aggregation function $h^{\text{o}}$. Specifically, $g^{\text{o}}$ takes a list of models $\gL$ as input and outputs the corresponding rankings for each index over all tasks, \ie, $\mR=g^{\text{o}}(\gL)$. 
    The rankings $\mR$ are fed into $h^{\text{o}}$, which outputs the final ranking $\vr^{\text{o}}=(r^{\text{o}}_1,r^{\text{o}}_2,\ldots,r^{\text{o}}_m)$, \ie, $\vr^{\text{o}}=h^{\text{o}}(\mR)$.
    \item We use $\textsc{rankdata}(\cdot)$ as the operator of getting rank.
\end{itemize}

More specifically, in cardinal benchmarks, an aggregated score is first calculated for each model, in most cases, by averaging the scores~\citep{wang2018glue,wang2019superglue}. 
For any candidate model list $\gL$, the final ranking is then calculated by sorting the average scores as follows,
\begin{equation}
\begin{split}
    \vr^{\text{c}}=h^{\text{c}}(\mS)
    &=\textsc{rankdata}\big((\bar{s}_1, \ldots, \bar{s}_m)\big)\,\text{,}
    \\
    \text{where}& \quad \bar{s}_i = \frac{1}{n} \sum_{j=1}^{n} s_{ij}\,\text{.}
    \label{eq:definition_rc}
\end{split}
\end{equation}
\noindent In contrast, most existing popular ordinal benchmarks in machine learning calculate the \emph{winning rate} for each model~\citep{liang2022holistic,Lee2023HolisticEO,bigcode-evaluation-harness}. 
For any candidate model list $\gL$, the winning rate for model $L_i$ represents the probability that its rank $r_{ik}$ is lower than $r_{jk}$ for a randomly selected opponent model~$L_j$ and task $T_k$.  
By referring to $\mathcal{I}(\cdot)$ as the indicator function, the final ranking is calculated as follows, 
\begin{equation}
\begin{split}
    \vr^{\text{o}}
    &=h^{\text{o}}(\mR)
    =\textsc{rankdata}\big((\bar{w}_1, \ldots, \bar{w}_m)\big)\,\text{,}
    \\
    \text{where}& \quad \bar{w}_i= \frac{1}{m} \sum_{j=1}^{m} w_{ij}\,\text{,}~~
    w_{ij}=\frac{1}{n}\sum_{k=1}^n \mathcal{I}(r_{ik}<r_{jk})\,\text{.}
    \label{eq:definition_ro}
\end{split}
\end{equation}

\paragraph{Arrow's Impossibility Theorem for Benchmarks}
Arrow's Impossibility Theorem, a cornerstone in social choice theory, posits that no system can flawlessly translate individual preferences into a group ranking~\citep{arrow1950difficulty,arrow2012social}.
Adapted to the case of multi-task benchmarks, the theorem says the following (proof in Appendix~\ref{app:proof}).
\begin{theorem}[Arrow's Impossibility Theorem for Benchmarks]
No ordinal benchmark $f^{\text{o}}$ can fulfill the following conditions simultaneously:
\begin{enumerate}
[itemsep=0mm, topsep=0mm]
    \item \textbf{Non-Dictatorship:} 
    There is no task $T_i$ such that, for any $\gL$ and any index pair $(x, y)$ , when $r_{xi} < r_{yi}$, then $r^{\text{o}}_{x} < r^{\text{o}}_{y}$.
    \item \textbf{Pareto Efficiency:} 
    For any $L$ and any index pair $(x, y)$, if $r_{xi} < r_{yi}$ for every task $T_i \in \gT$, then $r^{\text{o}}_{x} < r^{\text{o}}_{y}$.
    \item \textbf{Independence of Irrelevant Alternatives (IIA):} 
    Let $\gL$ and $\gL'$ be any two lists of models.
    For any index pair $(x, y)$, if $x$ and $y$ have the same relative order in $g^{\text{o}}(\gL)$ and $g^{\text{o}}(\gL')$ for all tasks, then $x$ and $y$ have the same relative order in $f^{\text{o}}(\gL)$ and $f^{\text{o}}(\gL')$.
    \item \textbf{Universality:} 
    The benchmark has at least three tasks.
    The benchmark has as domain all finite lists with at least three models.
    The scoring function $g^{\text{o}}$ has full range over all logically possible values for $\mR$.
    The aggregation function $h^{\text{o}}$ has full domain over all logically possible values for $\mR$.
\end{enumerate}
\label{thm:arrow}
\end{theorem}

For the ordinal benchmarks introduced in \eqref{eq:definition_ro}, the \emph{IIA} condition is especially problematic since introducing a new model can perturb the winning rate of existing models, and as a result, change the aggregated ranking of existing models.
For example, assume there are three candidate models $\gL=(L_1, L_2, L_3)$ and nine tasks $\gT=(T_1, T_2, \ldots, T_9)$, and the rankings $\mR$ are as follows,
\begin{itemize}[itemsep=0mm, topsep=0mm]
    \item for any task in $\{T_i\}_{i=1}^{4}$, $r_{1i}$ < $r_{2i}$ < $r_{3i}$,
    \item for any task in $\{T_i\}_{i=5}^{7}$, $r_{2i}$ < $r_{3i}$ < $r_{1i}$,
    \item for any task in $\{T_i\}_{i=8}^{9}$, $r_{3i}$ < $r_{1i}$ < $r_{2i}$.
\end{itemize}
The winning rates are $\bar{w}_1=10/27$, $\bar{w}_2=10/27$, $\bar{w}_3=7/27$, so we have $\vr^{\text{o}}_1=\vr^{\text{o}}_2<\vr^{\text{o}}_3$.
Now we add one extra candidate model and get $\gL'=(L_1, L_2, L_3, L_4)$. 
The rankings $\mR'$ are as follows,
\begin{itemize}[itemsep=0mm, topsep=0mm]
    \item for any task in $\{T_i\}_{i=1}^{4}$, $r'_{1i}$ < $r'_{2i}$ < $r'_{4i}$ < $r'_{3i}$,
    \item for any task in $\{T_i\}_{i=5}^{7}$, $r'_{2i}$ < $r'_{4i}$ < $r'_{3i}$ < $r'_{1i}$,
    \item for any task in $\{T_i\}_{i=8}^{9}$, $r'_{3i}$ < $r'_{1i}$ < $r'_{2i}$ < $r'_{4i}$.
\end{itemize}
Then the winning rates are $\bar{w}'_1=17/36$, $\bar{w}'_2=19/36$, , $\bar{w}'_3=9/36$ , so we have $\vr^{\text{o}\prime}_2<\vr^{\text{o}\prime}_1<\vr^{\text{o}\prime}_3$.
Note that the relative ranking among $\{L_1, L_2, L_3\}$ does not change over the nine tasks, but the final ranking has been different.

While Arrow's Impossibility Theorem is mainly concerned with ordinal voting systems, criticisms extend to cardinal systems as well.
The main concern lies in the interpersonal comparability between voters~\citep{Drakopoulos1989TheHP}.
The validity of interpersonal comparison has been challenged as transforming any particular scale for individual preferences has been widely recognized as arbitrary~\citep{Sen2017CollectiveCA,arrow2012social}.
In the context of cardinal benchmarks, the scale discrepancies among tasks could result in a situation where the aggregate performance disproportionately reflects the score of a single task, 
thereby distorting the benchmark's intent to represent all tasks effectively~\citep{colombo2022What,himmi2023More} and thus violating \emph{Non-Dictatorship}. 
As a result, outliers or skewed distributions can significantly influence the final ranking, undermining the reliability of the cardinal benchmark assessments.
Even if the scores are similar in scale across tasks, the underlying difficulty of each task may differ, \emph{i.e.}, improvements in one task are easier to come by than in another. 
Consequently, a cardinal benchmark would then reward improvements in the easier task more than in the harder task, leading to discrepancies in how improvements are valued.

Although this discussion suggests potential issues in multi-task benchmarks that are informed by Arrow's theorem, we have yet to establish quantitative metrics that can gauge the severity of these issues within existing benchmarks. The next section will provide such quantitative metrics.

\section{{Diversity} and {Sensitivity} in Multi-Task Benchmarks}
Inspired by Arrow's impossibility theorem, in this section, we propose two quantitative measurements for multi-task benchmarks, \emph{diversity} and \emph{sensitivity}. 
\emph{Diversity} is used to measure the ranking disagreement among all tasks, while \emph{sensitivity} measures how vulnerable the final ranking of the benchmarks is toward irrelevant changes that do not change the relative performance of models.

\subsection{{Diversity} in Multi-Task Benchmarks.}
\label{sec:diversity}
Let $\gL=(L_1, L_2,\ldots,L_m)$ contain all models in $\gM$ without duplicates, \ie, $m=|\gM|$, we define the \emph{diversity} with reversed Kendall's coefficient of concordance $W$, which is to assess the disagreement among model rankings on each task as follows,
\begin{equation}
\begin{split}
    W&= 1 - {12 \Sigma}/{(n^2(m^3-m))}\,\text{,}\\
    \text{where}\quad 
    \Sigma =& \sum_{i=1}^{m} (\bar{r}_i -  \tilde{r})^2\text{,}~~ 
    \tilde{r} = \sum_{i=1}^{m} r_i\text{,}~~
    \bar{r}_i = \sum_{j=1}^{n} r_{ij}\text{.}
    \label{eq:definition_w}
\end{split}
\end{equation}
$W=0$ means all model rankings are the same across all tasks, while $W=1$  means random or highly varied rankings.
For example, if the benchmark is composed of only one task, repeating multiple times, then the \emph{diversity} would be zero.

The definition of \emph{diversity} is inspired by the \emph{Universality} condition in Theorem~\ref{thm:arrow}, which indicates that a benchmark should accommodate all possible values for the rank matrix $\mR$, meaning that any configuration of model ranks across the tasks in $\gT$ should be feasible. 
This condition can be trivially satisfied when there is only one task in $\gT$ by rearranging the models in $\gL$.
However, it becomes challenging in a multi-task scenario, particularly if the tasks share high correlations in their evaluations of models. 
For instance, if all tasks in $\gT$ are merely replicas of a single task, the situation will never arise where the ranking vectors $\vr_i$ and $\vr_j$ differ, as such, not all values of $\mR$ are possible—thereby violating the \emph{Universality} condition.

As directly verifying \emph{Universality} is intractable, we use \emph{diversity} as an approximate. 
\emph{Diversity} quantifies the degree of alignment or discordance between rankings of different tasks over $\gM$.
A lower \emph{diversity} score indicates a strong inter-task correlation with similar rankings being produced across tasks, which could potentially impair \emph{Universality}, as it restricts the possible values that $\mR$ can take. 
For example, \emph{diversity} being zero means that all tasks are the same, and \emph{Universality} will be violated.
Conversely, a higher \emph{diversity} represents a stronger disagreement between tasks regarding model rankings, paving the way for more possible $\mR$ scenarios and thus aligning more closely with the tenet of \emph{Universality}. 
For example, \emph{diversity} being one means that the rankings of all tasks are random, and thus \emph{Universality} will hold.

\subsection{{Sensitivity} in Multi-Task Benchmarks}
\label{sec:sensitivity}
\emph{Sensitivity} is based on the desideratum from Arrow's theorem about the independence of irrelevant changes, restated below.
\begin{property}[Independence of Irrelevant Changes]
    The aggregated final ranking should not be altered by irrelevant changes on tasks that do not modify the relative performance of models.
\end{property}
\noindent Intuitively speaking, our measure of \emph{sensitivity} captures the degree to which a benchmark responds to irrelevant changes.
In particular, high \emph{sensitivity} implies that the desideratum of independence of irrelevant changes is strongly violated.

The definition of \emph{sensitivity} is different in the case of ordinal and cardinal benchmarks. 
Both definitions make use of Kendall's $\tau$ coefficient that we define next. 
For any model list $\gL$, Kendall's $\tau$ coefficient measures the distance between any two model rankings $\vr$ and $\vr'$, as follows,
\begin{align}
    \tau = \frac{\text{number of disconcordant pairs}}{\binom{m}{2}}\,\text{,}
    \label{eq:defintion_tau}
\end{align}
\noindent where a pair of models $L_i$ and $L_j$ is said to be concordant in $\vr$ and $\vr'$ if both $r_{i}>r_{j}$ and $r'_{i}>r'_{j}$ hold or both $r_{i}<r_{j}$ and $r'_{i}<r'_{j}$ hold; otherwise, this pair is considered as disconcordant.
One intuitive explanation for the number of disconcordant pairs is to count the number of times one has to cross lines when connecting matching data points from one ranking to another.
Here we have normalized $\tau$ into $[0, 1]$, so that
$\tau=0$ means that the two ranks are exactly the same, while $\tau=1$ means they are opposite to each other.
We primarily use $\tau$ as the measurement for ranking distance in our formulation, but we also report max rank change (\textit{MRC}) to provide a more intuitive measurement for the ranking distance in our experiments. 
For any two model rankings $\bm{r}$ and $\vr'$, \textit{MRC} is defined as follows,
\begin{align}
    \textit{MRC} = \max_{i \in \{1, 2, \ldots, m\}} \frac{|r_i - r'_i|}{m-1}\,\text{.}
    \label{eq:definition_mrc}
\end{align}
\noindent $\textit{MRC}=0$ means there is no ranking change, while $\textit{MRC}=1$ indicates the maximum possible fluctuation in rankings.
Next, we will define two kinds of irrelevant changes for cardinal and ordinal benchmarks, respectively. 

\paragraph{{Sensitivity} in cardinal benchmarks.}
For cardinal benchmarks, the \emph{sensitivity} is defined based on \emph{label noise injection} on tasks as the irrelevant change. 
Specifically, the injection of label noise to a task could well preserve the relative performances of models and should not change a task's intrinsic nature.
Therefore, \emph{sensitivity} aims to quantify the robustness of benchmark rankings to these task-equivalent manipulations in scores.
This concept is also loosely analogous to the 
\emph{Non-dictatorship} principle, which prohibits any single voter (in this case, a task and its scoring) from imposing an undue influence on the outcome.
In the most extreme scenario, randomizing all labels for a particular task is equivalent to excluding that task from the benchmark. 
Changes in the aggregated ranking brought by such manipulation can thus reveal the level of influence that the task has.
More significant fluctuations imply a greater impact of the task on overall rankings, suggesting that it plays a vital role in the benchmark, while minimal changes suggest that the task's influence is negligible.

Specifically, let $\gL=(L_1, L_2,\ldots,L_m)$ contain all models in $\gM$ without duplicates, \ie, $m=|\gM|$, we define \textit{sensitivity} by injecting different portions of label noise in each task, and calculating the largest ranking distance after injection, as follows,
\begin{align}
    \max_{\bm{\alpha}\in [\epsilon,1]^n} \quad
    &\tau(\vr^{\text{c}}, \vr' )\,
    \label{eq:cardinal_obj}
    \\ 
    \text{s.t.}~~~~~&
    \vr'=\textsc{rankdata}\big((\bar{s}'_1, \bar{s}'_2, \ldots, \bar{s}'_m)\big)\,\text{,}
    \label{eq:updated_score_cardinal}
    \\
    & 
    \bar{s}'_i=\sum_{j=1}^{n} \big( \alpha_j s_{ij}+(1-\alpha_j) p_{j} \big)\,\text{,} 
    \label{eq:updated_score}
\end{align}
\noindent where the original ranking $\vr^{\text{c}}$ and scores $s_{ij}$ are defined in Section~\ref{sec:perspective}.
$\bm{\alpha}=(\alpha_1, \alpha_2, \ldots, \alpha_n)$ defines the proportions of preserved examples whose labels are unchanged for each task.
The label noise are injected by randomly substituting $(1-\alpha_j)$ portion of examples' labels into a random one.
As a result, the corresponding score for these examples would be $p_j$, which refers to the performance under random label substitution.
For example, if the task $T_j$ is binary classification and its score refers to accuracy, the $p_j$ refers to $1/2$.
In practice, the specific value of $p_j$ does not have any impact on the ranking $\vr'$ as the sum $\sum_{j=1}^{n} (1-\alpha_j)p_j$ does not depend on models.
$\epsilon \in (0,1)$ is a hyper-parameter that defines the minimal preserving proportion of unchanged examples for each task.
It is worth noting that $\vr'$ will not change if we multiply a positive constant with $\bm{\alpha}$, so we could always keep the maximum value in $\bm{\alpha}$ as one by multiplying $\bm{\alpha}$ with $1 / max(\{\alpha_i\}_{i=1}^{n})$, which means that there is always at least one task with no noise.

\paragraph{{Sensitivity} in ordinal benchmarks.}
For ordinal benchmarks, the definition for \textit{sensitivity} is built upon \emph{irrelevant model addition} as the irrelevant change.
This is inspired by the \emph{IIA} condition in Arrow's Impossibility Theorem, which demands that the addition of a new model (an “irrelevant alternative” with respect to existing comparisons) should not change the relative ranking order of the models already being considered.
If a benchmark's model rankings are dramatically affected every time a new model is introduced into the competition, it indicates a high \emph{sensitivity}.
A low \emph{sensitivity} in ordinal benchmarks assures us that the relative rankings are stable and that the benchmark can handle the introduction of new models without disrupting the existing rankings, consistent with the IIA condition. 
This resilience is essential, as it means that the benchmark's evaluations are reliable and reflective of each model's true performance relative to its peers.

Specifically, let $\gL=(L_1, L_2, \ldots, L_m)$ be a list of models, and $\gL^{\text{C}}=(L_{m+1}, L_{m+2}, \ldots, L_{m+l})$ represent the complement model list, \ie, $m+l=|\gM|$.
Then the \emph{sensitivity} for ordinal benchmarks is defined as the largest ranking distance after adding a subset of these extra candidate models into comparison.
\begin{align}
&~~~~~~~~~~~~~~~~~~~
\max_{\bm{\beta} \in \{0,1\}^l}
\tau(\vr^{\text{o}}, \vr' )\,
\label{eq:ordinal_obj}
\\
\text{s.t.}~~&
\vr'
=\textsc{rankdata}\big((\bar{w}'_1, \bar{w}'_2, \ldots, \bar{w}'_m)\big)\,
\text{,} 
\label{eq:updated_score_ordinal}
\\
& 
\bar{w}'_i= \frac{1}{m+||\bm{\beta}||_1} \sum_{j=1}^{m} w_{ij} + \sum_{j=1}^{l} \beta_j w_{i(m+j)}\,\text{,}
\label{eq:updated_win_rate}
\end{align}
\noindent where we use $\bm{\beta} \in \{0,1\}^l$ as the indicator irrelevant model selection, where $\beta_j=1$ means $M_{m+j}$ is selected and $\beta_j=0$ means not-selected.
As a result, $||\bm{\beta}||_1$ refers to the number of selected models from $\gL^{\text{C}}$ as irrelevant models.

Assuming \emph{IIA} in Theorem~\ref{thm:arrow} holds, for any model list $\gL$, after appending a list of irrelevant models (as indicated by $\bm{\beta}$), the relative ranking among models in $\gL$ should not change.
In practice, we simply calculate \emph{sensitivity} by selecting the top-$20\%$ models in the existing benchmark as $\gL$ and the rest $80\%$ models as $\gL^{\text{C}}$. 
If \emph{IIA} holds, then \emph{sensitivity} should be zero.
On the other hand, we note that our \emph{sensitivity} is a lower bound for \emph{IIA}, which says that \emph{IIA} could still not hold even if \emph{sensitivity} is zero.
This limit mainly comes from the setting where we only consider the top-$20\%$ models, which is inspired by real-world scenarios where most people only care about top models.
Note that, in our paper, we only focus on ordinal benchmarks that aggregate the final ranking by calculating the winning rate as in \eqref{eq:definition_ro}, which satisfies \emph{Non-Dictatorship} and \emph{Pareto Efficiency} by design. 
For ordinal benchmarks with other aggregation methods, one should take all \emph{Non-Dictatorship}, \emph{Pareto Efficiency}, and \emph{IIA} into consideration for \emph{sensitivity}.

\begin{figure}
\begin{minipage}[t]{0.48\textwidth}
\begin{algorithm}[H]
   \caption{Sensitivity for Cardinal Benchmarks}
   \label{alg:cardinal}
\begin{algorithmic}[1]
   \STATE {\bfseries Input:} scores $\{s_{ij}\}_{i\in[1,m], j\in[1,n]}$ for models in $\gL$, $\epsilon$, $\lambda$, number of optimization $T$
   \STATE Calculate $\vr^{\text{c}}$ based on \eqref{eq:definition_rc}
   \STATE Initialize the parameter $\vtheta \in R^{n}$ randomly
   \FOR{$t=1$ {\bfseries to} $T$}
       \STATE $\bm{\alpha} = Sigmoid(\vtheta) + {\epsilon}/{(1 - \epsilon)} $
       \STATE $\bm{\alpha} = \alpha / ||\alpha||_1$
       \STATE Calculate updated score $\{\bar{s}'_i\}_{i=1}^{m}$ as \eqref{eq:updated_score}
       \STATE Calculate relaxed loss $\ell^{\text{c}}$ based on \eqref{eq:cardinal_obj_relax}
       \STATE Gradient descent on $\vtheta$ based on $\ell^{\text{c}}$
   \ENDFOR
   \STATE $\bm{\alpha} = Sigmoid(\vtheta) + {\epsilon}/{(1 - \epsilon)} $
   \STATE $\bm{\alpha} = \alpha / max(\bm{\alpha})$
   \STATE Calculate $\vr'$ based on \eqref{eq:updated_score_cardinal}
   \STATE {\bfseries Output:} $\tau(\vr^{\text{c}}, \vr')$
\end{algorithmic}
\end{algorithm}
\end{minipage}
\begin{minipage}[t]{0.48\textwidth}
\begin{algorithm}[H]
   \caption{Sensitivity for Ordinal Benchmarks}
   \label{alg:ordinal}
\begin{algorithmic}[1]
   \STATE {\bfseries Input:} winning rates $\{w_{ij}\}_{i\in[1,m+l], j\in[1,m+l]}$ for $\gL$ and $\gL^{\text{C}}$, $\lambda$, number of optimization $T$
   \STATE Calculate $\vr^{\text{o}}$ based on \eqref{eq:definition_ro}
   \STATE Initialize the parameter $\vtheta \in R^{n}$ randomly
   \FOR{$t=1$ {\bfseries to} $T$}
       \STATE $\bm{q}_{\bm{\beta}} = Sigmoid(\vtheta)$
       \STATE $\bm{\beta} \sim  Bernoulli(\bm{q}_{\bm{\beta}})$
       \STATE $\bm{\beta} = \bm{\beta} + \bm{q}_{\bm{\beta}} - \bm{q}_{\bm{\beta}}.detach()$
       \STATE Calculate updated win rate $\{\bar{w}'_i\}_{i=1}^{m}$ as \eqref{eq:updated_win_rate}
       \STATE Calculate relaxed loss $\ell^{\text{o}}$ based on \eqref{eq:ordinal_obj_relax}
       \STATE Gradient descent on $\vtheta$ based on $\ell^{\text{c}}$
   \ENDFOR
   \STATE $\bm{\beta}= (Sigmoid(\vtheta) > 0.5).int()$
   \STATE Calculate $\vr'$ based on \eqref{eq:updated_score_ordinal}
   \STATE {\bfseries Output:} $\tau(\vr^{\text{o}}, \vr')$
\end{algorithmic}
\end{algorithm}
\end{minipage}
\end{figure}

\paragraph{Relaxation of {sensitivity}.}
The main challenge for solving \eqref{eq:cardinal_obj} and \eqref{eq:ordinal_obj} lies in the non-differentiable nature of the operator $\textsc{rankdata}(\cdot)$ and $\tau(\cdot, \cdot)$.
Thus we propose to relax the ranking distance to a continuous objective as follows,
\begin{align}
    \ell^{c} &= \sum_{i=1}^{m} \sum_{j =1}^{m} 
    \bigg(
    \mathcal{I}(r^{\text{c}}_i < r^{\text{c}}_j)
    \max(\bar{s}'_i - \bar{s}'_j, -\lambda)
    \bigg)
    \label{eq:cardinal_obj_relax}
    \\
    \ell^{o} &= \sum_{i=1}^{m} \sum_{j =1}^{m} 
    \bigg(
    \mathcal{I}(r^{\text{o}}_i < r^{\text{o}}_j)
    \max(\bar{w}'_i - \bar{w}'_j, -\lambda) 
    \bigg)
    \label{eq:ordinal_obj_relax}
\end{align}
\noindent where $\mathcal{I}(\cdot)$ is the indicator function, and $\lambda \geq 0$ is a hyperparameter.
If the optimal point of \eqref{eq:cardinal_obj_relax} could be achieved, for any $r^{\text{c}}_i<r^{\text{c}}_j$, 
we have  
$r'_i>r'_j$
because 
$s'_i<s'_j$.
As a result, the original objectives in \eqref{eq:cardinal_obj} would also achieve the optimal solution as $\tau(\vr^{\text{c}}, \vr')=1$ based on \eqref{eq:defintion_tau}.
The same applies to ordinal benchmarks with \eqref{eq:ordinal_obj} and \ref{eq:ordinal_obj_relax}.

The algorithms for calculating \emph{sensitivity} for cardinal and ordinal benchmarks could be seen in Algorithm~\ref{alg:cardinal} and \ref{alg:ordinal}.
For cardinal \emph{sensitivity} calculation in Algorithm~\ref{alg:cardinal}, we normalize the sum of $\bm{\alpha}$ as one in line 5-6 during optimization, or otherwise the loss could be minimized by setting $\bm{\alpha} = \bm{0}$.
For ordinal \emph{sensitivity} calculation in the algorithm~\ref{alg:ordinal}, in order to handle the optimization challenge brought by the discrete nature of $\bm{\beta}$, we sample it from a Bernoulli distribution with probability $\bm{q}_{\bm{\beta}}$ modeled by $\vtheta$ as shown in line 6.
The straight through technique~\citep{Jang2016CategoricalRW,Bengio2013EstimatingOP} is used to handle the gradients on $\vtheta$.
Due to the potential approximation errors and optimization challenges, the calculated ranking distances by both algorithms are the lower bound of the true values.

\begin{table}[t!]
\caption{Summary of Benchmarks}
\label{tab:benchmark_summary}
\centering
\resizebox{0.66\textwidth}{!}{
\begin{tabular}{c|c|c|c}
\toprule
\midrule
\textbf{Type} & \textbf{Benchmark} & \textbf{No. of Tasks} & \textbf{No. of Models} \\
\midrule
\multirow{10}{*}{Cardinal} & \texttt{GLUE}                &  9  & 87  \\
& \texttt{SuperGLUE}           & 8  & 28  \\
& \texttt{BIG-Bench-Hard}      & 27 & 107 \\
& \texttt{MTEB}                & 56 & 83  \\
& \texttt{OpenLLM}             & 6  & 100 \\
& \texttt{MMLU}                & 57 & 100 \\
& \texttt{VTAB}                & 19 & 16  \\
& \texttt{ImageNet}                & 20 & 112  \\
& \texttt{Random}                & 100 & 100  \\
& \texttt{Constant}                & 100 & 100  \\
\midrule
\multirow{13}{*}{Ordinal} & \texttt{BigCode}             & 3  & 41  \\
& \texttt{HELM-accuracy}       & 16 & 67  \\
& \texttt{HELM-fairness}       & 14 & 67  \\
& \texttt{HELM-robustness}     & 14 & 67  \\
& \texttt{HEIM-alignment-auto} & 40 & 26  \\
& \texttt{HEIM-quality-auto}   & 12 & 26  \\
& \texttt{HEIM-aesthetics-auto} & 60 & 26  \\
& \texttt{HEIM-alignment-human} & 23 & 26  \\
& \texttt{HEIM-nudity}         & 20 & 26  \\
& \texttt{HEIM-quality-human}  & 7  & 26  \\
& \texttt{HEIM-aesthetics-human} & 18 & 26  \\
& \texttt{Random}                & 100 & 1000  \\
& \texttt{Constant}                & 100 & 1000  \\
\midrule
\bottomrule
\end{tabular}}
\end{table}

\section{Experiments on Cardinal Benchmarks}
In this section, we present the \emph{diversity} and \emph{sensitivity} to label noise injection for seven cardinal benchmarks.

\paragraph{Experiment setup} For our experiment, we have collected seven widely-used benchmarks for our experiments, \texttt{GLUE}~\citep{wang2018glue}, \texttt{SuperGLUE}~\citep{wang2019superglue}, \texttt{MTEB}~\citep{muennighoff2022mteb}, \texttt{BigBenchHard}~\citep{Suzgun2022ChallengingBT}, \texttt{MMLU}~\citep{hendrycks2020measuring},  \texttt{OpenLLM}~\citep{open-llm-leaderboard,eval-harness} and \texttt{VTAB}~\citep{Zhai2019ALS}.
To provide a better understanding of the \emph{diversity} and \emph{sensitivity} spectrums, we further introduce three additional ``baseline'' benchmarks,
\texttt{Constant}, \texttt{Random} and \texttt{ImageNet}.
The \texttt{Constant} benchmark features a single task where the scores for 100 different models are randomly determined, and the task has been duplicated 100 times. 
The \texttt{Random} benchmark assigns random scores to all 100 models across all 100 tasks.
The \texttt{ImageNet} benchmark is based on the validation set of the ILSVRC-2012 challenge~\citep{Deng2009ImageNetAL}.
We divide its 1,000 classes into 20 equally-sized subsets at random, with each subset functioning as a distinct task. 
We conducted evaluations on 112 models that had been pretrained on ImageNet and were sourced from the TorchVision~\citep{torchvision2016}. 
The average performance across these 20 tasks corresponds to the original accuracy metric, thus ensuring that the final rankings are consistent with those derived from the original accuracy measures.
More details of all benchmarks are in Table~\ref{tab:benchmark_summary} and Appendix~\ref{app:benchmark_detail}.

\emph{Diversity} and \emph{sensitivity} scores are computed for each benchmark based on \eqref{eq:definition_w} and Algorithm~\ref{alg:cardinal}.
For both measures, all models are used for calculation, \ie, $\gL$ contains all models in the leaderboard.
The only exceptions are \texttt{OpenLLM} and \texttt{MTEB}, where we focus on the top-100 models out of thousands of candidates to mitigate the influence of less reliable ones.
For the \emph{sensitivity} calculation in each benchmark, we set minimal preserving portion $\epsilon=\min\{0.01, \mathrm{std}_{\min}/\mathrm{std}_{\max}\}$, where $\mathrm{std}_{\min}$ and $\mathrm{std}_{\max}$ refer to the smallest and largest standard deviations across all tasks in the benchmark respectively.
If all tasks have the same standard deviation, this will ensure that at least 1\% of the data remains unaltered by label noise in each task.
However, if there is variability in the standard deviations across tasks, $\epsilon$ will be adjusted based on the standard deviation.
This adjustment prevents scenarios where a single task with a significantly larger standard deviation disproportionately influences the sensitivity calculation.
$\lambda$ is set as 0.0 and the number of gradient descent $T$ is 1000.
Results for \texttt{Constant}, \texttt{Random} and \texttt{ImageNet} are averaged over five random trials.

\begin{figure}[t!]
    \centering
    \includegraphics[width=0.46\textwidth]{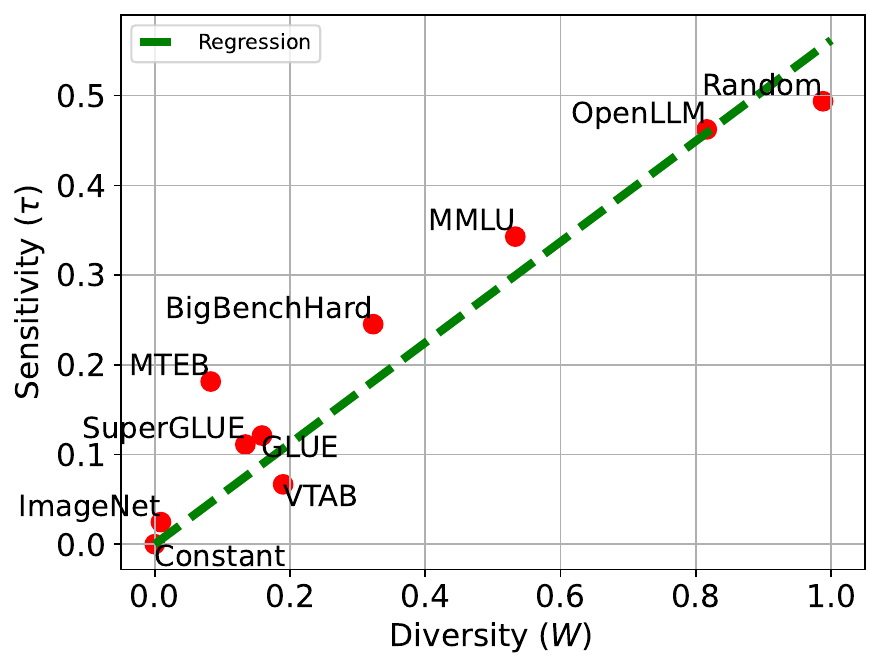}
    \includegraphics[width=0.46\textwidth]{figures/cardinal_trade_off_MRC.pdf}
    \caption{
    The $x$-axis indicates the \textit{diversity} of model rankings across tasks, evaluated by the Kendall's $W$ coefficient.
    The $y$-axis represents the \textit{sensitivity} of the final model ranking to different portions of label noise across tasks.
    The ranking change is measured by both Kendall's $\tau$ (top) and \textit{MRC} (bottom).
    The green curve is by linear regression on all points without fitting intercept.
    }
    \label{fig:cardinal_trade_off}
\end{figure}

\paragraph{Experiment results}
The results are presented in Figure~\ref{fig:cardinal_trade_off}.
A strong positive correlation between \emph{diversity} and \emph{sensitivity} can be observed, with Pearson correlation of 0.96 and 0.77 for top and bottom figures, respectively.
A larger \emph{diversity} always comes at the cost of high \emph{sensitivity} to label noise injection.
\texttt{Constant} is the most stable benchmark, while \texttt{Random} achieves the highest \emph{diversity}.
All real-world multi-task benchmarks roughly strike a trade-off comparable with the linear interpolation between \texttt{Random} and \texttt{Constant}. 

Both \emph{diversity} and \emph{sensitivity} vary a lot across different benchmarks.
For example, \texttt{OpenLLM} achieves the second largest \emph{diversity} ($W=0.82$) and suffers from a high \emph{sensitivity} ($\tau=0.54,\textit{MRC}=0.86$).
In contrast, benchmarks like \texttt{GLUE} ($W=0.16,\tau=0.11, \textit{MRC}=0.72$) and \texttt{SuperGLUE} ($W=0.13,\tau=0.12, \textit{MRC}=0.33$) demonstrate far lower \emph{diversity} and \emph{sensitivity}. 
The underlying reason can be two-fold.
First, the tasks within \texttt{GLUE} and \texttt{SuperGLUE} are more similar to each other by definition. 
For example, \texttt{GLUE} primarily consists of NLI and text classification tasks.
In contrast, tasks within \texttt{OpenLLM} are more messy, including commonsense inference and reasoning, math problems, science questions, etc.
Second, the candidate models in \texttt{OpenLLM} are also more noisy due to the relatively lower entry barrier. 
In contrast, there are a lot of restrictions for participant models to get presented in the leaderboard in \texttt{GLUE} and \texttt{SuperGLUE}, and thus enjoy a lower chance of having outlier candidate models in the leaderboard.

The results of \texttt{ImageNet} serve as a sanity check of our choice for the minimal preserving portion, denoted by $\epsilon$.
\texttt{ImageNet} has been one of the most influential single-task benchmarks in the field of machine learning, and its evaluation results have been widely regarded as a solid measure of progress in model development~\citep{Dwork2015TheRH, Tsipras2020FromIT,koch2021Reduced}.
Despite being split into 20 tasks, our \texttt{ImageNet} essentially parallels the original single-task benchmark in terms of model rankings.
The experiment results show that \texttt{ImageNet} achieves the second lowest \emph{sensitivity}, only slightly higher than \texttt{Constant}.
It demonstrates that rankings of a high-quality benchmark remain stable even when subjected to significant label noise. 
Such robustness emphasizes the importance of resisting noise interference for benchmarks and validates our choice for $\epsilon$.

\begin{figure}[t!]
    \centering
    \includegraphics[width=0.46\textwidth]{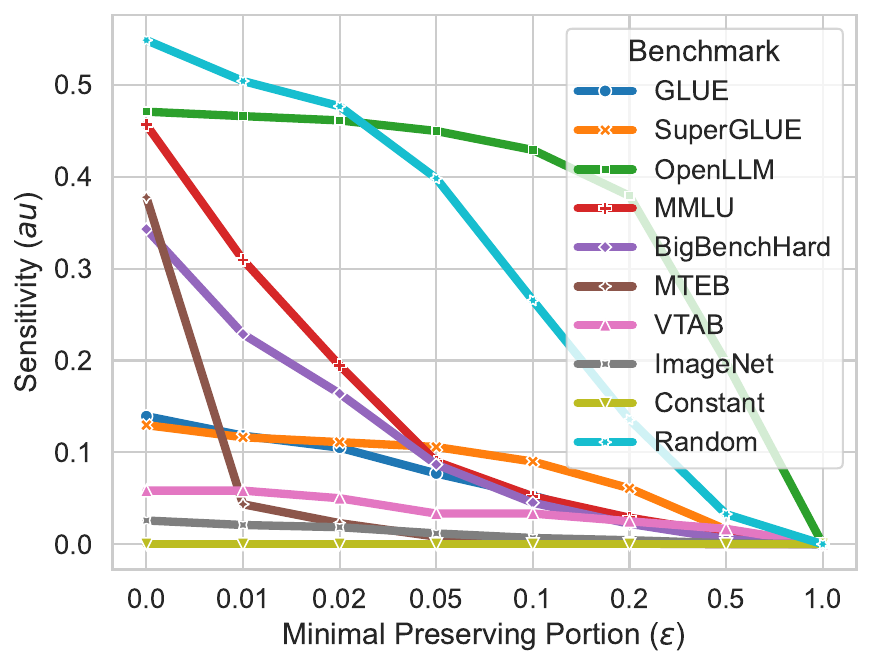}
    \includegraphics[width=0.46\textwidth]{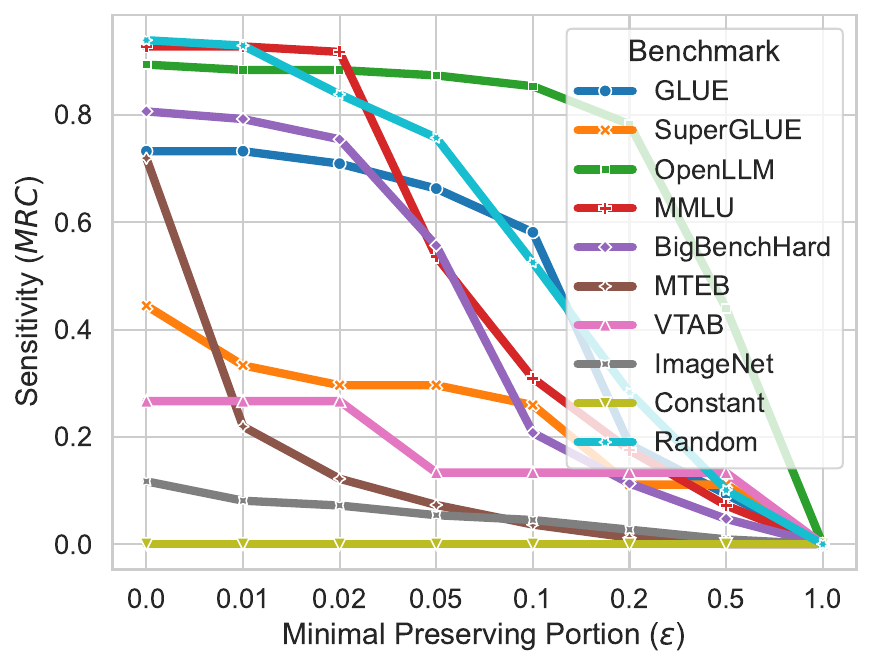}
    \caption{Sensitivity of cardinal benchmarks as a function of the minimal preserving ratio $\epsilon$. 
    $x$-axis refers to the minimal preserving portion of unchanged examples, $\epsilon$, as stated in \eqref{eq:cardinal_obj}.
    The $y$-axis refers to \emph{sensitivity} measured by $\tau$ (top) and \textit{MRC} (bottom).
    }
    \label{fig:cardinal_minvalue}
\end{figure}

To delve deeper into benchmark \emph{sensitivity} concerning the minimal preserving portion $\epsilon$, Figure~\ref{fig:cardinal_minvalue} plots the \emph{sensitivity} across varying $\epsilon$ values. 
When preserving 10\% of the data ($\epsilon=0.1$), the \textit{MRC} for all non-baseline benchmarks ranges from 0.18 to 0.71, indicating a non-trivial ranking change. 
Notably, \texttt{OpenLLM} maintains a $\tau$ of 0.13 and \textit{MRC} of 0.45 even at $\epsilon=0.5$, underscoring its pronounced volatility.

\section{Experiments on Ordinal Benchmarks}
In this section, we present \emph{diversity} and \emph{sensitivity} to irrelevant models over eleven ordinal benchmarks.

\paragraph{Experiment setup}
Our selected benchmarks for experiments consist of \texttt{BigCode}~\citep{bigcode-evaluation-harness}, three benchmarks from \texttt{HELM}~\citep{liang2022holistic}, and seven benchmarks from \texttt{HEIM}~\citep{Lee2023HolisticEO}.
The original rankings for all these benchmarks are based on the winning rate, as defined in \eqref{eq:definition_ro}.
We excluded any benchmarks that suffered from a lot of missing values or that showcased an undifferentiated scoring pattern among different models as these complicate the calculation of the winning rate. 
The remaining benchmarks are \texttt{HELM-accuracy}, \texttt{HELM-fairness}, \texttt{HELM-robustness},
 \texttt{HEIM-alignment-auto}, \texttt{HEIM-quality-auto}, \texttt{HEIM-aesthetics-auto}, \texttt{HEIM-alig-}
\texttt{nment-human}, \texttt{HEIM-nudity}, \texttt{HEIM-quality-human}, \texttt{HEIM-aesthetics-human}.
The statistics can be seen in Table~\ref{tab:benchmark_summary}, and more details are in Appendix~\ref{app:benchmark_detail}.
Similar to cardinal benchmarks, we also add \texttt{Constant} and \texttt{Random} benchmarks, with 100 tasks and 1000 models for each.

For both \emph{diversity}, all models in each benchmark are used for calculation, \ie, $\gL$ contains all models in the leaderboard.
To calculate \emph{sensitivity}, we use the original top-$20\%$ models in the leaderboard as $\gL$, and the ranking calculated only based on them is referred to as $\vr^{\text{o}}$.
Then we use the rest models as $\gL^{\text{C}}$, use Algorithm~\ref{alg:ordinal} to select a subset from $\gL^{\text{C}}$ as irrelevant models to alter the rankings of $\vr^{\text{o}}$.
The only exception is \texttt{Random}, where we use the top-10 models as $\gL$ and use the rest as $\gL^{\text{C}}$, in order to simulate the scenario with infinite potential irrelevant models. 
$\lambda$ is set as 0.01 and the number of gradient descent $T$ is 100.
For the calculation of \emph{diversity}, we impute the missing scores with a KNN imputer as the rankings for each task must be of the same dimension for calculating Kendall's $W$.

\begin{figure}[t!]
    \centering
    \includegraphics[width=0.46\textwidth]{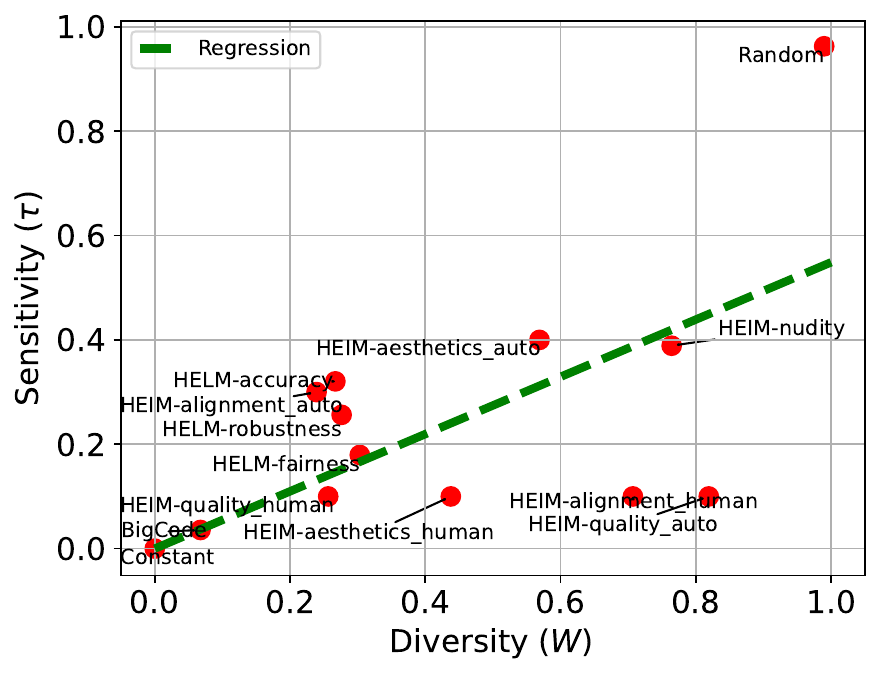}
    \includegraphics[width=0.46\textwidth]{figures/ordinal_trade_off_MRC.pdf}
    \caption{
    The $x$-axis indicates the \emph{diversity} of model rankings across tasks, evaluated by the reversed Kendall's $W$ coefficient, where $W=0$ denotes uniformity in rankings, while $W=1$ means random or highly varied rankings across tasks. 
    The $y$-axis represents the \emph{sensitivity} to irrelevant candidate models addition, measured by the Kendall's $\tau$ (top) and \textit{MRC} (bottom).
    The green curve is by linear regression on all points without fitting intercept.
    }
    \label{fig:ordinal_trade_off}
\end{figure}

\paragraph{Experiment results}
The results are shown in Figure~\ref{fig:ordinal_trade_off}, where we plot the \emph{diversity} and \emph{sensitivity} towards additional irrelevant alternative models.
The green curve acquired by fitting all points, demonstrates that there is a strong correlation between \emph{diversity} and \emph{sensitivity}.
The Pearson correlation is 0.61 and 0.50 for both figures.
The lower Pearson correlation (compared to cardinal benchmarks) and the observed deviation from the regression curve could be attributed to the missing values in \texttt{HEIM} and \texttt{HELM}-based benchmarks. 
The KNN imputation method is used to impute values so that \emph{diversity} could be calculated, but this also might lead to inaccuracies in the \emph{diversity} estimation. 

Several benchmarks exhibit significant \emph{sensitivity}. 
For instance, a notable change in ranking is observed with the \texttt{HEIM-aesthetic-auto} benchmark, where the \textit{MRC} reaches as high as 0.5 and $\tau$ reaches 0.4.
Moreover, over half of these benchmarks exhibit an MRC of at least 0.5, which highlights their vulnerability to the inclusion of irrelevant models.
The dependency of rankings on the selection of candidate models casts doubts on the reliability of the evaluation outcomes of these benchmarks.

One outlier in both plots is the \texttt{Random} benchmark, which is relatively far away from the regression curve.
This anomaly can be attributed to the assumption that the \texttt{Random} benchmark contemplates a nearly infinite array of irrelevant models for selection. 
Consequently, it allows for greater flexibility in altering the rankings of the existing models. 
This also suggests that, as the number of candidate models increases over time, the aggregated final rankings could be more unstable.
\section{Conclusion}
In this work, we examine multi-task benchmarks through the lens of social choice theory. 
Our exploration starts by applying Arrow's impossibility theorem on ordinal benchmarks, suggesting that there may be intrinsic limitations for multi-task benchmarks. But Arrow's theorem is neither quantitative, nor does it apply to cardinal benchmarks. We therefore develop two key measures of multi-task benchmarks---task diversity and stability to irrelevant changes---that we argue stand in tension with one another. Our empirical investigations on seven cardinal benchmarks and eleven ordinal benchmarks yield insights about the inherent trade-off between the two proposed measures.
Furthermore, our analysis reveals significant sensitivity issues in several popular benchmarks, calling into question the validity of evaluation outcomes derived from these benchmarks. 

\section{Acknowledgement}
We would like to thank 
Joachim Baumann, 
Andr\`e Cruz, 
Ricardo Dominguez-Olmedo,
Florian E. Dorner,
and
Celestine Mendler-Dünner
for helpful discussions and/or feedback on draft versions of this work.

\bibliography{references}
\bibliographystyle{plainnat}

\newpage
\appendix
\onecolumn

\section{Benchmark Details}
\label{app:benchmark_detail}
We use the following benchmarks for our experiment.
The cardinal benchmarks are as follows,
\begin{itemize}
[itemsep=0mm, topsep=0mm]
\item The \texttt{GLUE} benchmark is designed to evaluate natural language understanding models using 9 tasks that cover fundamental linguistic abilities such as sentiment analysis, entailment, and similarity prediction. 
There are 87 candidate models.
The leaderboard can be found in \url{https://gluebenchmark.com/leaderboard}.
\item \texttt{SuperGLUE}, as an extension of GLUE, consists of more demanding tasks aimed at assessing deeper linguistic comprehension and commonsense reasoning, spanning 8 tasks. 
There are 28 candidate models.
The leaderboard can be found in \url{https://super.gluebenchmark.com/leaderboard}.
\item \texttt{BIG-Bench-Hard}, a subset of the larger BIG-Bench, zeroes in on 27 specifically challenging tasks to test models on complex reasoning and understanding nuanced language. 
There are 107 candidate models. The leaderboard is found in \url{https://opencompass.org.cn/dataset-detail/BBH}.
\item \texttt{MTEB} is designed to extensively evaluate text embeddings, including 56 datasets across 7 different tasks and covering 112 languages to seek a universal text embedding method. 
There are 83 candidate models. The leaderboard is found in \url{https://huggingface.co/spaces/mteb/leaderboard}.
As the original leaderboard reports the weighted average based on the number of datasets within each task, we simply repeat each task correspondingly in our experiment and use direct averaging.
\item \texttt{OpenLLM} leaderboard evaluates open language models with 6 key benchmarks applied in the EleutherAI Language Model Evaluation Harness, involving various tasks related to reasoning, general knowledge, and truthfulness in both zero-shot and few-shot frameworks. Only the top 100 candidate models are used for our experiment.
The leaderboard can be found in \url{https://huggingface.co/spaces/HuggingFaceH4/open_llm_leaderboard}.
\item The \texttt{MMLU} benchmark offers a large-scale, multidisciplinary evaluation with a focus on academic knowledge, including 57 different subjects. Only the top 100 candidate models are used for our experiment.
The leaderboard can be found in \url{https://huggingface.co/spaces/HuggingFaceH4/open_llm_leaderboard}.
\item \texttt{VTAB}, short for Visual Task Adaptation Benchmark, is a suite designed to evaluate the versatility and generalizability of visual representations by measuring performance across 19 diverse classification tasks without using evaluation datasets during pre-training. 
There are 16 candidate models.
The leaderboard can be found in \url{https://google-research.github.io/task_adaptation/benchmark}.
\end{itemize}
The ordinal benchmarks are as follows,
\begin{itemize}
[itemsep=0mm, topsep=0mm]
\item \texttt{BigCode} is designed to test the abilities of code generation models by posing complex coding challenges in 3 programming languages. 
There are 41 models in the benchmark.
The leaderboard can be found in \url{https://huggingface.co/spaces/bigcode/bigcode-models-leaderboard}.
\item \texttt{HELM} evaluates language models across 8 scenarios, where each scenario corresponds to a benchmark with multiple tasks.
We excluded any benchmarks that suffered from a lot of missing values or that showcased an undifferentiated scoring pattern among different models, as these would result in tied results frequently, which subsequently complicates the calculation of winning rates.
Three benchmarks have remained: 
\texttt{HELM-accuracy} (16 tasks),
\texttt{HELM-fairness} (14 tasks), and
\texttt{HELM-robustness} (14 tasks).
There are 67 candidate models in each benchmark.
The leaderboard can be found in \url{https://crfm.stanford.edu/helm/lite/latest/#/leaderboard}.
\item \texttt{HEIM} is tailored to scrutinize the performance of text-to-image models across ten dimensions such as creativity, equity, and language coverage, each of which forms a benchmark with multiple tasks. 
We excluded any benchmarks that suffered from a lot of missing values or that showcased an undifferentiated scoring pattern among different models, as these would result in tied results frequently, which subsequently complicates the calculation of winning rates.
Seven benchmarks have remained:
\texttt{HEIM-alignment-auto} (40 tasks),
\texttt{HEIM-quality-auto} (12 tasks),
\texttt{HEIM-aesthetics-auto} (60 tasks),
\texttt{HEIM-alignment-human} (23 tasks),
\texttt{HEIM-nudity} (20 tasks),
\texttt{HEIM-quality-human} (7 tasks),
\texttt{HEIM-aesthetics-human} (18 tasks).
There are 26 candidate models in each benchmark.
The leaderboard can be found in \url{https://crfm.stanford.edu/heim/latest/?group=core_scenarios}.

\end{itemize}
Each of these benchmarks collectively aims to provide a comprehensive platform to test the limits and versatility of machine learning models from multiple aspects.
The statistics could be seen in Table~\ref{tab:benchmark_summary}.

\newpage
\section{Proof of Arrow's Impossibility Theorem for Benchmarks}
\label{app:proof}

We include a proof of Arrow's result in our notation for the sake of completeness.

\paragraph{Notation.}
We first restate the notation, as follows:
\begin{itemize}
[itemsep=0mm, topsep=0mm]
    \item $\gT=(T_1, T_2, \ldots, T_n)$ represents the list of all $n$ tasks in the benchmark, analogous to voters.
    \item $\gM$ refers to the set of all potential candidate models that could be evaluated by the benchmark. 
    \item Let $\gL=(L_1, L_2, \ldots, L_m)$ be any non-empty list of candidate models with $m$ models, where $L_i \in \mathcal{M}$ for any $i$.
    \item For any $\gL$, we define $s_{ij}$ as the score for the $i$-th model in $\gL$ in task $T_j$.
    For simplicity, we abuse the notations and use $\vs_j=(s_{1j}, s_{2j}, \ldots, s_{mj})$ as scores in any task $T_j$, and $\mS=(\vs_1, \vs_2, \ldots, \vs_n)$ as scores over all tasks.
    \item For any $\gL$, we define $r_{ij}$ as the rank for the $i$-th model $\gL$ in task $T_j$ \emph{w.r.t.} $\gL$.
    For simplicity, we abuse the notations and use $\vr_j=(r_{1j}, r_{2j}, \ldots, r_{mj})$ as ranks in any task $T_j$, and $\mR=(\vr_1, \vr_2, \ldots, \vr_n)$ as ranks over all tasks.
    \item A cardinal benchmark is defined as a function $f^{\text{c}}=h^{\text{c}} \circ g^{\text{c}}$, which is composed of the scoring function $g^{\text{c}}$ and the aggregation function $h^{\text{c}}$. Specifically, $g^{\text{c}}$ takes a list of models $\gL$ as input and outputs the corresponding scores for each index over all tasks, \ie, $\mS=g^{\text{c}}(\gL)$. 
    The scores $\mS$ are fed into 
    $h^{\text{c}}$, which outputs the final ranking $\vr^{\text{c}}=(r^{\text{c}}_1,r^{\text{c}}_2,\ldots,r^{\text{c}}_m)$, \ie, $\vr^{\text{c}}=h^{\text{c}}(\mS)$.
    \item An ordinal benchmark is defined as a function $f^{\text{o}}=h^{\text{o}} \circ g^{\text{o}}$, which is composed of the scoring function $g^{\text{o}}$ and the aggregation function $h^{\text{o}}$. Specifically, $g^{\text{o}}$ takes a list of models $\gL$ as input and outputs the corresponding rankings for each index over all tasks, \ie, $\mR=g^{\text{o}}(\gL)$. 
    The rankings $\mR$ are fed into $h^{\text{o}}$, which outputs the final ranking $\vr^{\text{o}}=(r^{\text{o}}_1,r^{\text{o}}_2,\ldots,r^{\text{o}}_m)$, \ie, $\vr^{\text{o}}=h^{\text{o}}(\mR)$.
    \item We use $\textsc{rankdata}(\cdot)$ as the operator of getting rank.
\end{itemize}

\paragraph{Arrow's Impossibility Theorem for Benchmarks}
We present Arrow's Impossibility Theorem for benchmarks as follows,
\begin{theorem}[Arrow's Impossibility Theorem for Benchmarks]
No ordinal benchmark $f^{\text{o}}$ can fulfill the following conditions simultaneously:
\begin{enumerate}
[itemsep=0mm, topsep=0mm]
    \item \textbf{Non-Dictatorship:} 
    There is no task $T_i$ such that, for any $\gL$ and any index pair $(x, y)$ , when $r_{xi} < r_{yi}$, then $r^{\text{o}}_{x} < r^{\text{o}}_{y}$.
    \item \textbf{Pareto Efficiency:} 
    For any $L$ and any index pair $(x, y)$, if $r_{xi} < r_{yi}$ for every task $T_i \in \gT$, then $r^{\text{o}}_{x} < r^{\text{o}}_{y}$.
    \item \textbf{Independence of Irrelevant Alternatives (IIA):} 
    Let $\gL$ and $\gL'$ be any two lists of models.
    For any index pair $(x, y)$, if $x$ and $y$ have the same relative order in $g^{\text{o}}(\gL)$ and $g^{\text{o}}(\gL')$ for all tasks, then $x$ and $y$ have the same relative order in $f^{\text{o}}(\gL)$ and $f^{\text{o}}(\gL')$.
    \item \textbf{Universality:} 
    The benchmark has at least three tasks.
    The benchmark has as domain all finite lists with at least three models.
    The scoring function $g^{\text{o}}$ has full range over all logically possible values for $\mR$.
    The aggregation function $h^{\text{o}}$ has full domain over all logically possible values for $\mR$.
\end{enumerate}
\label{app_thm:arrow}
\end{theorem}

\paragraph{Supporting Lemmas}
    To prove the Theorem~\ref{app_thm:arrow}\footnote{The proof is largely the same as the original Arrow's Theorem in \url{https://shorturl.at/bdlI0}.}, we first define decisive coalitions and present two supporting lemmas:
    \begin{itemize}
[itemsep=0mm, topsep=0mm]
        \item A subset of tasks $\gG \subset \gT$ is a coalition.
        \item A coalition $\gG$ is decisive over an index pair $(x, y)$ if and only if, for any $\mL$, when $r_{xi}<r_{yi}$ for every $T_i \in \gG$, then $r^{\text{o}}_{x} < r^{\text{o}}_{y}$.
        \item A coalition $\gG$ is decisive if and only if it is decisive over all ordered pairs.
        \item A coalition $\gG$ is decisive over an index pair $(x, y)$ if and only if, for any $\mL$, when $r_{xi}<r_{yi}$ for every $T_i \in \gG$ and $r_{xj}>r_{yj}$ for every $T_j \in (\gT - \gG)$, then $r^{\text{o}}_{x} < r^{\text{o}}_{y}$.
    \end{itemize}
    \begin{lemma}[Field Expansion Lemma]
        For a benchmark that satisfies \emph{Pareto Efficiency}, \emph{IIA} and \emph{Universality}, 
        if a coalition $\gG$ is weakly decisive over index pair $(x,y)$ for some $x \neq y$, then it is decisive.
    \end{lemma}
    \begin{proof}
        Assume $\gG$ is weakly decisive over $(x,y)$.
        Let $z$ be any index distinct from $x$ and $y$.
        Find a $\gL$ such that $r_{xi}<r_{yi}<r_{zi}$ for every task $T_i \in \gG$, and $r_{yj}<r_{xj}$ and $r_{yj}<r_{zj}$ for every task $T_j \in (\gT - \gG)$.
        Note that there is no need to specify the relationship between $r_{xj}$ and $r_{zj}$ for $T_j \in (\gT - \gG)$.
        By \emph{Pareto Efficiency}, we have $r^{\text{o}}(y)<r^{\text{o}}(z)$.
        By weak decisiveness of $\gG$ over $(x, y)$, we have $r^{\text{o}}(x)<r^{\text{o}}(y)$.
        Thus we have $r^{\text{o}}(z)<r^{\text{o}}(y)$ for $\gL$.
        By \emph{IIA}, every $\gL'$ which shares the same relative order for $(x, z)$, \ie, $r_{xi}<r_{zi}$ for every task $T_i \in \gG$, should have $r^{\text{o}}(z)<r^{\text{o}}(y)$.
        Therefore, $\gG$ is decisive over $(x,z)$.
        Similarly, we could show $\gG$ is also decisive over $(y,z)$.
        Therefore, we prove that $\gG$ is decisive for all index pairs in $\{x,y,z\}$. 
        Iterating the above process, we could prove that $\gG$ is decisive for all index pairs in $\{1, 2, \ldots, m\}$, and thus the proof is complete.
    \end{proof}
    \begin{lemma}[Group Contraction Lemma]
        For a benchmark that satisfies \emph{Pareto Efficiency}, \emph{IIA} and \emph{Universality}, 
        if a coalition $\gG$ is decisive, and has at least two tasks, then it has a proper subset that is also decisive.
    \end{lemma}
    \begin{proof}
        Assume $\gG$ is decisive and has at least two tasks. Partition $\gG$ into $\gG_1$ and $\gG_2$.
        Fix distinct indices $x, y, z$. 
        Find a $\gL$ such that 
        \begin{align}
            r_{xi} < r_{yi} < r_{zi}&~~~~\text{if}~~~~T_i \in \gG_1 \\
            r_{zi} < r_{xj} < r_{yj}&~~~~\text{if}~~~~T_j \in \gG_2 \\
            r_{yk} < r_{zk} < r_{xk}&~~~~\text{if}~~~~T_k \in (\gT-\gG)
        \end{align}
        Since $\gG$ is decisive, we have $r^{\text{o}}_x<r^{\text{o}}_y$.
        So at least one is true between 
        $r^{\text{o}}_x<r^{\text{o}}_z$ and
        $r^{\text{o}}_z<r^{\text{o}}_y$.
        If $r^{\text{o}}_x<r^{\text{o}}_z$, then $\gG_1$ is weakly decisive over $(x, z)$.
        If $r^{\text{o}}_z<r^{\text{o}}_y$, then $\gG_2$ is weakly decisive over $(z, y)$.
        Now apply the Field Expansion Lemma. 
        By iterating the process, the lemma is proved.
    \end{proof}

\paragraph{Proof of Arrow's Impossibility Theorem for Benchmarks}
\begin{proof}
    By \emph{Pareto Efficiency}, the entire set of tasks $\gT$ is decisive, thus by \emph{Group Contraction Lemma}, there is a size-one decisive coalition — a dictator. 
    In other words, any benchmark that satisfies \emph{Pareto Efficiency}, \emph{IIA} and \emph{Universality} will violate  \emph{Non-Dictatorship}.
    Hence, the proof is complete.
\end{proof}

\newpage
\section{Additional Experimental Results on Diversity}

\begin{figure}[t!]
    \centering
    \includegraphics[width=0.46\textwidth]{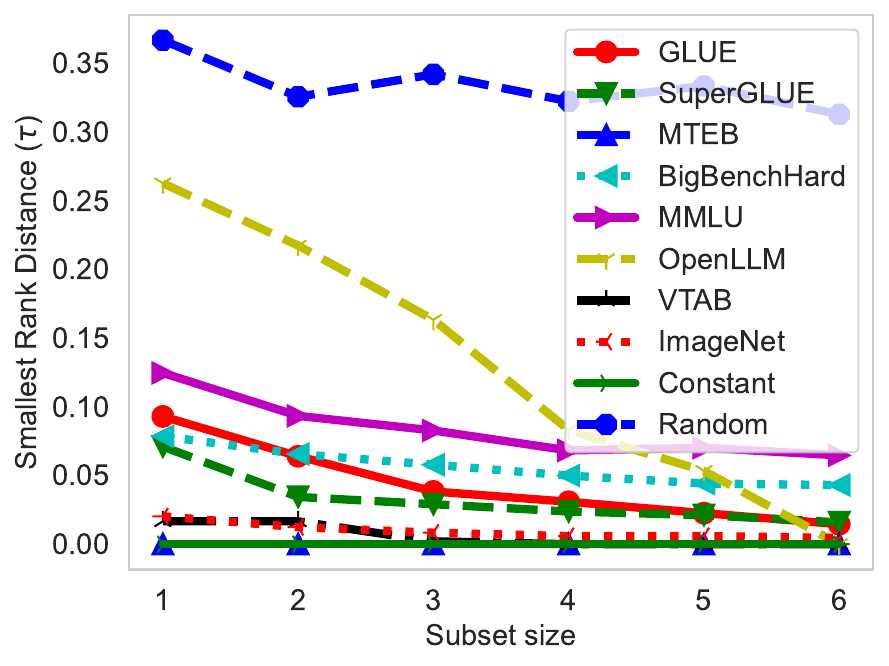}
    \includegraphics[width=0.46\textwidth]{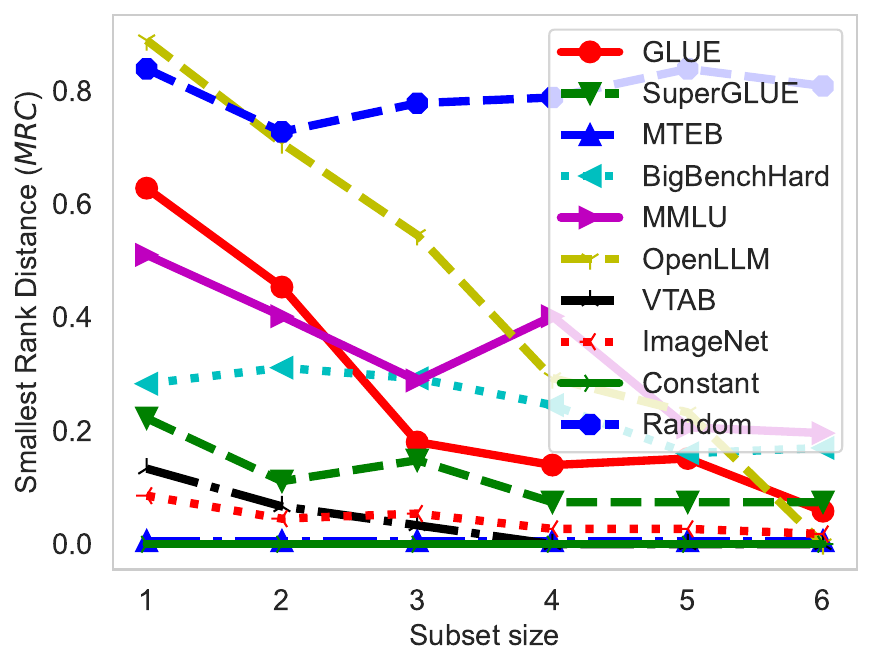}
    \includegraphics[width=0.46\textwidth]{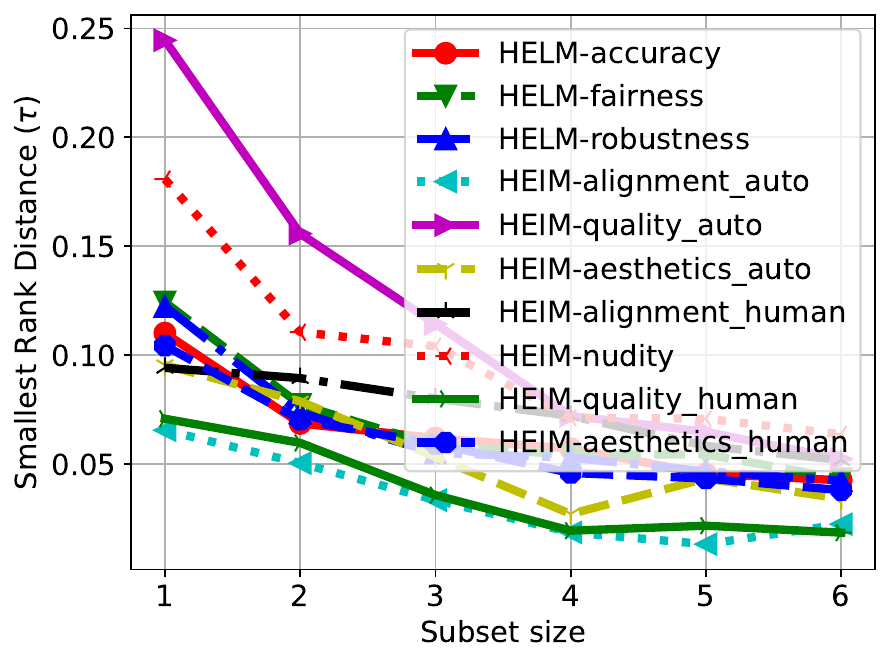}
    \includegraphics[width=0.46\textwidth]{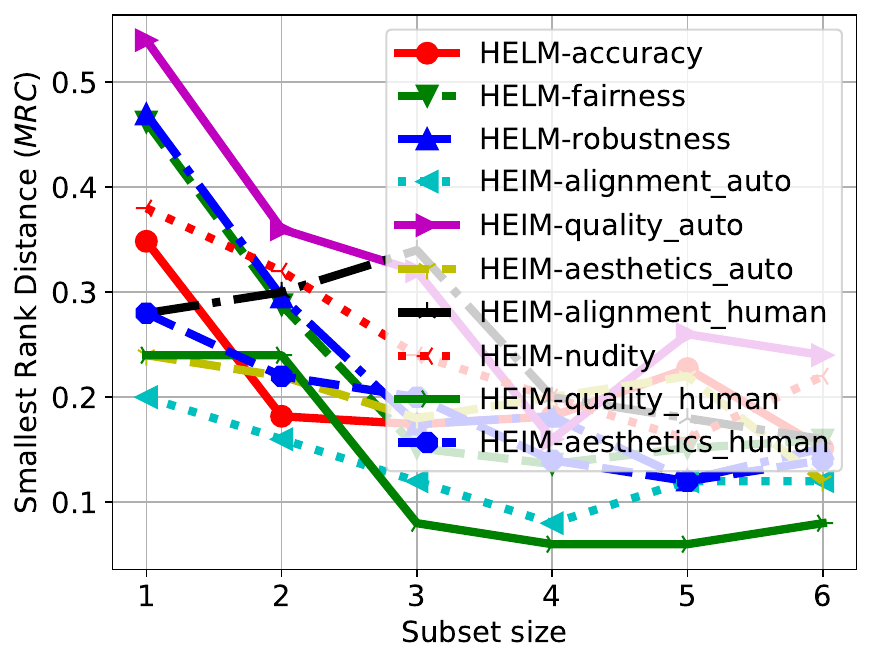}
    \caption{
        Smallest rank change by re-calculating the average score based on a subset of tasks.
        $x$-axis refers to how many tasks are selected in the subset. 
        By randomly sampling 1000 random subsets of the specific size, we report the smallest ranking distance from the original ranking in $y$-axis, in cardinal benchmarks (top) and ordinal benchmarks (bottom), measured by Kendall's $\tau$ (left) and $\textit{MRC}$ (right).
    }
    \label{fig:cardinal_subset_min}
\end{figure}

We further conduct an experiment, seeking to determine the minimum number of tasks necessary to obtain an approximation of the overall final ranking.
We examined subsets of tasks ranging from sizes one to six, randomly sampling these subsets for 1000 times and identifying which offered a ranking closest to the overall ranking.
We exclude \texttt{BigCode} for this experiment as it only contains three tasks.
The outcomes are illustrated in Figure~\ref{fig:cardinal_subset_min}.
Intriguingly, our findings align with the \emph{diversity} present within the benchmarks. 
For instance, \texttt{OpenLLM} and \texttt{HEIM-quality-auto}, which display the greatest \emph{diversity} (except for \texttt{Random}) in Figure~\ref{fig:cardinal_trade_off} and \ref{fig:ordinal_trade_off}, also requires the largest number of tasks to arrive at a ranking proximate to the overall ranking. 
Conversely, benchmarks exhibiting less \emph{diversity}, such as \texttt{VTAB} and \texttt{HEIM-quality-human}, require fewer tasks to replicate the overall ranking. 
This suggests that benchmarks with lower \emph{diversity} might contain more redundant tasks that do not significantly contribute to the overall ranking.

While this experiment offers valuable insights into the connection between \emph{diversity} and the minimum number of tasks needed to approximate the overall ranking, it is important to acknowledge its limitations. 
The results can be influenced by the number of tasks in the benchmarks, potentially skewing the findings. 
For example, for a benchmark with three tasks, the maximum number of tasks to approximate the overall ranking is always three, no matter how large \emph{diversity} of the benchmark is.
To mitigate this issue, one potential approach could be to consider the ratio of tasks rather than absolute numbers. 
However, this will introduce another challenge where tasks could be duplicated within a benchmark to artificially reduce the minimal subset ratio required to replicate the full ranking. 
Despite these limitations, the experiment provides an intuitive understanding of how \emph{diversity} correlates with the minimum subset size necessary for ranking recovery. 
We recognize the need for further exploration in future research endeavors.

\end{document}